\documentclass[letterpaper, 10 pt, conference]{ieeeconf}  

\IEEEoverridecommandlockouts                              

\usepackage[utf8]{inputenc}

\usepackage{amsmath,amssymb,amsfonts,amsthm}
\usepackage{bm}
\usepackage{url}
\usepackage{graphicx,graphics}
\usepackage{multirow}
\usepackage{booktabs}
\usepackage{multirow}
\usepackage[ruled,linesnumbered]{algorithm2e}
\usepackage{textcomp}
\usepackage{tabularx}
\usepackage{relsize}
\usepackage[flushleft]{threeparttable}
\usepackage{cite}
\usepackage[table,dvipsnames]{xcolor}
\usepackage{colortbl}
\usepackage{enumerate}
\let \labelindent \relax 
\usepackage[inline]{enumitem}
\usepackage{acronym}
\usepackage{siunitx}
\usepackage{subfig}
\usepackage{comment}

\makeatletter
\let\NAT@parse\undefined
\makeatother
\usepackage{hyperref}  
\usepackage[capitalise]{cleveref}
\hypersetup{colorlinks=true, citecolor=black, linkcolor=Red, urlcolor=black}

\makeatletter
\newcommand*\titleheader[1]{\begingroup\gdef\@titleheader{#1}\let\footnote=\thanks\endgroup}
\AtBeginDocument{
  \let\st@red@title\@title
  \def\@title{
  \begin{flushleft}
    \vspace{-2.15em}
    \bgroup\normalfont\small\@titleheader\par\egroup
    \vspace{-18pt}\par\noindent\rule{\textwidth}{0.1pt}
    \end{flushleft}
    \vskip0.7em\st@red@title
        }
}

\makeatother

\title{\LARGE \bf
Tactile-Conditioned Diffusion Policy for \\ Force-Aware Robotic Manipulation
}

\titleheader{\textit{Preprint}}

\author{
    Erik Helmut$^{1}$, Niklas Funk$^{1}$, Tim Schneider$^{1}$, Cristiana de Farias$^{1}$, Jan Peters$^{1,2,3,4,5}$
    \thanks{Corresponding author: Erik Helmut. Email: erik@robot-learning.de.}
    \thanks{$^{1}$Department of Computer Science, Technical University of Darmstadt \quad $^{2}$German Research Center for AI (DFKI) \quad $^{3}$ Centre for Cognitive Science, Technical University of Darmstadt \quad $^{4}$ Hessian Center for Artificial Intelligence (Hessian.AI), Darmstadt \quad $^{5}$Robotics Institute Germany (RIG)}
\thanks{This work was supported by the German Federal Ministry of Education and Research (BMBF) and the French Research Agency, l’Agence Nationale de Recherche (ANR), through the project \emph{Aristotle} (ANR-21-FAI1-0009-01) and the EU’s Horizon Europe project ARISE (Grant no.: 101135959).}
\thanks{This work has been submitted to the IEEE for possible publication. Copyright may be transferred without notice, after which this version may no longer be accessible.}
}

\begin{document}\sloppy

\maketitle
\thispagestyle{empty}
\pagestyle{empty}

\bstctlcite{IEEEexample:BSTcontrol}

\newacro{fea}[FEA]{Finite Element Analysis}
\newacro{feats}[FEATS]{Finite Element Analysis for Tactile Sensing}
\newacro{mdp}[MDP]{Markov Decision Process}
\newacro{bc}[BC]{Behavioral Cloning}
\newacro{irl}[IRL]{Inverse Reinforcement Learning}
\newacro{ee}[EE]{End-Effector}
\newacro{ros}[ROS]{Robot Operating System}
\newacro{ddpm}[DDPM]{Denoising Diffusion Probabilistic Models}
\newacro{ddim}[DDIM]{Denoising Diffusion Implicit Models}
\newacro{ci}[CI]{Confidence Interval}
\newacro{ecdf}[ECDF]{Empirical Cumulative Distribution Function}
\newacro{umi}[UMI]{Universal Manipulation Interface}
\newacro{cnn}[CNN]{Convolutional Neural Network}
\newacro{film}[FiLM]{Feature-wise Linear Modulation}
\newacro{farm}[FARM]{Force-Aware Robotic Manipulation}
\begin{abstract}
Contact-rich manipulation depends on applying the correct grasp forces throughout the manipulation task, especially when handling fragile or deformable objects.
Most existing imitation learning approaches often treat visuotactile feedback only as an additional observation, leaving applied forces as an uncontrolled consequence of gripper commands.
In this work, we present \ac{farm}, an imitation learning framework that integrates high-dimensional tactile data to infer tactile-conditioned force signals, which in turn define a matching force-based action space.
We collect human demonstrations using a modified version of the handheld Universal Manipulation Interface (UMI) gripper that integrates a GelSight Mini visual tactile sensor.
For deploying the learned policies, we developed an actuated variant of the UMI gripper with geometry matching our handheld version.
During policy rollouts, the proposed \ac{farm} diffusion policy jointly predicts robot pose, grip width, and grip force.
\ac{farm} outperforms several baselines across three tasks with distinct force requirements—high-force, low-force, and dynamic force adaptation—demonstrating the advantages of its two key components: leveraging force-grounded, high-dimensional tactile observations and a force-based control space.
The codebase and design files are open-sourced and available at \url{https://tactile-farm.github.io}.
\end{abstract}

\section{Introduction}
\label{sec:intro}
Humans naturally regulate grasp forces through touch, applying just enough pressure to prevent an object from slipping \cite{johansson1984roles}, \cite{johansson1987signals}.
In robotics, the selection of an appropriate grasping force has long been recognized as a crucial issue~\cite{bicchi2000robotic}, especially when handling fragile or deformable objects, such as a fruit or an egg. 
In such cases, it is essential to employ the appropriate grasping force to minimize the risk of slippage or breakage.
Tactile sensing has therefore emerged as a key component of forceful manipulation, enabling slip detection and inference of shear and normal forces \cite{dahiya2010tactilesensingfromhumanstohumanoids}. 
Yet, despite its importance, effectively harnessing tactile sensing for direct force control in robots remains challenging.

One direction to address this challenge is imitation learning, which has gained traction as an efficient way of learning robotic manipulation by leveraging human demonstrations \cite{osa2018algorithmic}.
Recent works have incorporated tactile feedback into imitation learning approaches.
However, in most of these methods, tactile sensing is treated primarily as an additional observation modality, useful to get information for resolving visual occlusion or detecting contact state, but not as a signal that directly shapes the action space \cite{xue2025reactivediffusionpolicyslowfast}.
As a result, tactile feedback influences the policies only indirectly through its effect on the observation embedding, while the applied forces themselves remain an uncontrolled consequence of gripper commands.

\begin{figure}
    \centering
    \includegraphics[width=0.475\textwidth]{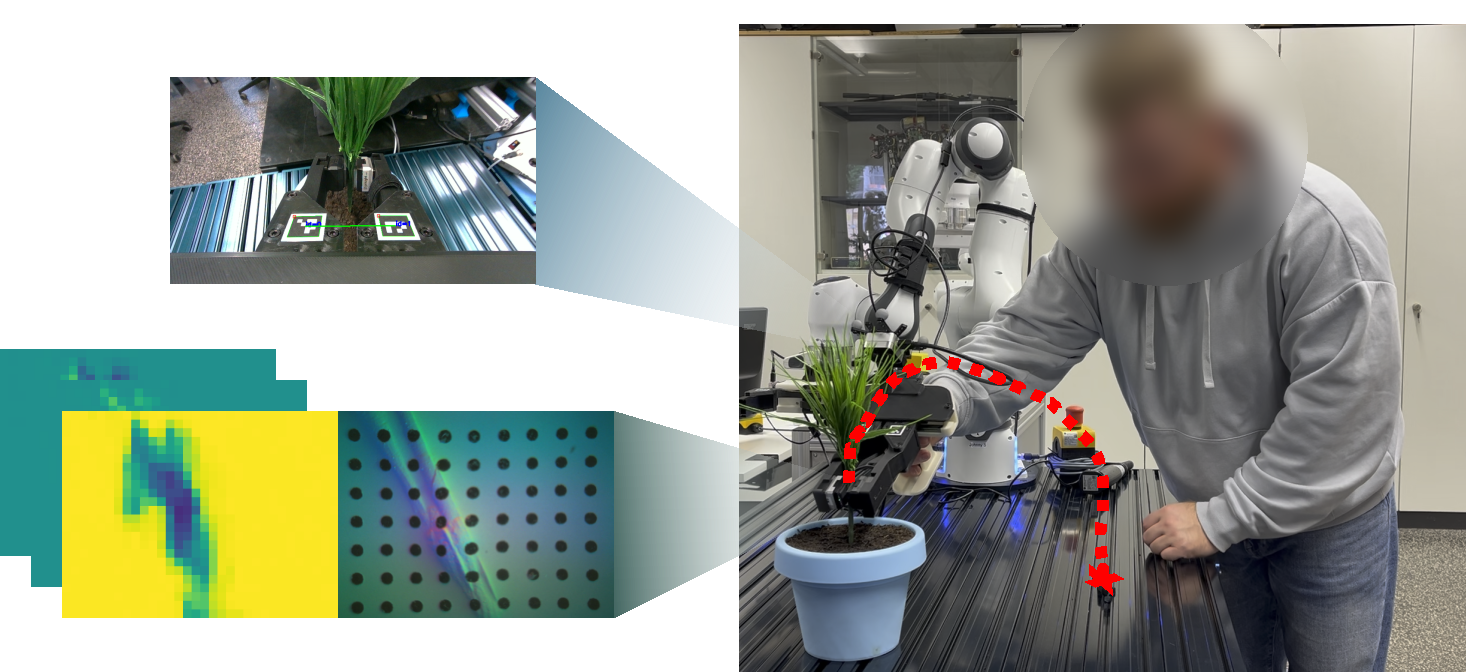}
    \caption{Data-collection setup using the adapted hand-held UMI gripper. Right: An expert performs a task using the adapted hand-held UMI gripper. Top left: In-hand RGB camera view with ArUco markers for grip width measurement. Bottom left: GelSight Mini tactile image and corresponding FEATS force estimates, visualizing the contact interaction and force distributions during demonstration.}
    \label{fig:data_collection}
\end{figure}

What remains largely missing is an imitation learning framework in which tactile feedback is not only perceived but also explicitly incorporated in the action space. 
Such a formulation would allow the policy to target and regulate the tactile interaction it intends to produce, rather than leaving contact forces as an uncontrolled side-effect of kinematic and gripper control.

In this work, we address this gap by introducing \acf{farm}, an imitation learning framework that integrates tactile feedback into the action space, while also leveraging a high-dimensional force-profile as an observation modality.
For this purpose, we leverage the GelSight Mini \cite{gelsightminionline2025}, a high-resolution vision-based tactile sensor, together with \acf{feats} \cite{helmut2025learningforcedistributionestimation} for estimating shear and normal contact forces.
The demonstrations are collected using an adapted hand-held Universal Manipulation Interface (UMI) gripper \cite{chi2024universal}, as shown in \cref{fig:data_collection}.
Using the expert demonstrations, we train a diffusion policy, which we execute on a Franka Research 3 robot with an actuated variant of the UMI gripper.
The \emph{Actuated UMI} gripper, which we introduce in this work, has a matching geometry with our hand-held UMI gripper, and thus functions as a drop-in replacement without the need for retargeting.
During execution, the gripper operates under a dual-mode control scheme: it alternates between position control of the grip width and closed-loop force control, depending on the presence of contact. This approach ensures stable and reliable performance in both free-space and contact conditions. 

The main contributions of this work are summarized as follows:

\begin{itemize}
    \item We present the \ac{farm} diffusion policy, which predicts robot pose, target grip width, and target grip force jointly, with forces represented both in the observation space and as an explicit action, yielding temporally consistent, force-aware action sequences.
    \item We design and build the \emph{Actuated UMI} gripper as an open-source platform, enabling direct transfer of demonstrations collected with the adapted hand-held UMI gripper equipped with a GelSight Mini sensor.
    \item We analyze and ablate the effects of different control modes and tactile representations on task success rates across three real-world tasks: plant insertion, grape picking, and screw tightening.
\end{itemize}

\section{Related Works}
\label{sec:relatedworks}
Tactile sensing is a key modality for advancing contact-rich robotic manipulation \cite{dahiya2010tactilesensingfromhumanstohumanoids, cutkosky1993manipulation, mandil2023tactile}, complementing vision by providing information about forces \cite{helmut2025learningforcedistributionestimation}, \cite{funk2023high}, \cite{sferrazza2019ground}, texture \cite{bohm2024matters}, \cite{manfredi2014natural}, and slip \cite{chen2018tactile}, \cite{yuan2015measurement}, \cite{funk2024evetaceventbasedopticaltactile}.
Indeed, tactile feedback is increasingly being integrated into learning-based manipulation frameworks \cite{funk2025importancetactilesensingimitation}.
In the context of visuotactile imitation learning, prior work can broadly be divided into two categories: approaches that use the tactile feedback to enrich the observation space, and approaches that aim to exploit tactile information more directly in shaping the learned actions.

\subsection{Incorporating Visuotactile Observations into Robotic Manipulation}

Works such as 3D-ViTac~\cite{huang20243dvitac}, GelFusion~\cite{jiang2025gelfusionenhancingroboticmanipulation}, and TactileAloha~\cite{aloha11063285} integrate tactile and visual signals into a unified latent representation, enabling policies to either overcome visual occlusion or to leverage contact information during manipulation. These, however, all rely on teleoperation for data collection.
Alternatively, works such as Ablett et al.~\cite{Ablett_2025} rely on data collected through kinesthetic teaching.
In this case, the kinesthetic data is converted back into robot actions through a tactile force matching objective that is active when replaying the trajectories using an impedance controller. This replay of the trajectories collected through kinesthetic teaching then yields the pose setpoints which are ultimately used for policy training.
This idea conceptually aligns with DexForce~\cite{chen2025dexforceextractingforceinformedactions}, which leverages contact forces, measured on a robotic hand with F/T sensors during kinesthetic demonstrations.
By converting the measured forces into force-informed position targets via an impedance controller, the robot can replay the demonstrations, yielding trajectories suitable for policy learning.
While all these methods highlight the utility of leveraging tactile information for policy learning, tactile feedback remains an auxiliary observation signal rather than a modality that is directly exploited within the controller.
Moreover, these methods also rely on demonstration data collection with the robot-in-the-loop either through kinesthetic teaching and subsequent force-informed trajectory replay~\cite{Ablett_2025}, \cite{chen2025dexforceextractingforceinformedactions}, or teleoperation~\cite{huang20243dvitac}, \cite{jiang2025gelfusionenhancingroboticmanipulation}, \cite{aloha11063285}. Such strategies inherently couple the demonstrations to the specific robot embodiment, which can limit generalization across platforms. By contrast, we adopt a robot-free data collection paradigm using an adapted hand-held UMI gripper that records visuotactile observations, thereby mitigating the embodiment gap.

In parallel, recent works have also adopted a robot-free data collection paradigm through custom visuotactile manipulation interfaces. 
MimicTouch~\cite{yu2025mimictouchleveragingmultimodalhuman} builds on non-parametric imitation learning, where tactile and audio embeddings together with the robot end-effector pose are matched against a demonstration library to retrieve a nearest-neighbor-based action prediction. Online residual reinforcement learning is then used to adapt the policies learned from human demonstrations for robotic execution.
ViTaMIn~\cite{liu2025vitaminlearningcontactrichtasks} also exploits a customized \acf{umi} gripper and proposes a multimodal representation learning strategy to obtain a tactile representation that captures essential contact properties, such as the object's in-hand pose and gripper's deformation.
FreeTacMan~\cite{wu2025freetacmanrobotfreevisuotactiledata} demonstrate that tactile features combined with visual observations improve success in contact-rich manipulation tasks compared to vision-only approaches.
However, compared to this work, in all of these systems, the policy ultimately outputs joint or Cartesian commands, leaving tactile feedback as an indirect signal that is not directly exploited for representing the actions.

\subsection{Tactile Sensing for Action Representation}
Instead of passively conditioning the policy on tactile signals, some works use these signals to regulate actions, marking an important step toward force-aware control. For instance, Xu et al.~\cite{xue2025reactivediffusionpolicyslowfast} developed the TactAR teleoperation system to collect demonstration data with real-time visual tactile/force feedback. Building on this data, the authors propose a Reactive Diffusion Policy, where a latent diffusion policy predicts action chunks in latent space at low frequency, while the fast asymmetric tokenizer refines these latent actions at high frequency using real-time tactile feedback, effectively acting as a learned impedance controller.
While their method focuses on controlling the forces exerted by the robot arm, our approach specifically controls the grip force, where grip force and gripper width are themselves actions predicted by the policy. 
By embedding forces directly into the action representation, our method provides finer-grained control at the contact interface. Furthermore, while the reactive diffusion policy separates planning and reactive refinement into two subsystems, our single diffusion policy jointly predicts the robot pose, grasp width, and grasp force trajectories, yielding a simpler and more transparent architecture.

Closely related to our work, Adeniji et al.~\cite{adeniji2025feelforcecontactdrivenlearning} introduce the Feel-the-Force framework, which also incorporates tactile sensing directly into the action space. Using human demonstrations collected with a tactile glove, their policy predicts gripper end-effector poses together with grasping forces, which are then executed through a PD force controller. This approach, however, relies on a calibrated setup and a reset alignment between the human hand and robot gripper, as well as manual annotation of semantic keypoints to initialize scene representations. Moreover, their execution is constrained by binarized gripper states and requires the force controller to converge before the robot advances to the next action, which slows execution and limits adaptability. By contrast, our method learns directly from robot-embodied demonstrations through a hand-held gripper, avoiding cross-domain retargeting. Instead of a binary gripper state, we predict continuous grip width and target forces, enabling smoother and more precise control. A dual-mode controller scheme decides when to regulate grip width versus grip force, allowing execution to proceed without waiting for force convergence and enabling adaptation to real-time tactile feedback.

Together, these works demonstrate the promise of incorporating tactile sensing into the action space. 
Yet, they either do not control the grasp force at all or use a restrictive, synchronized control scheme, in which the agent cannot control the target gripper width.
Our approach provides advances in this direction by directly coupling continuous grip force control with diffusion policy learning from robot-embodied tactile data, enabling a continuous, adaptive, and contact-aware manipulation.

\section{Method}
\label{ch:methods}
Here, we introduce \acf{farm}, an imitation learning framework to learn tactile-informed manipulation policies from demonstration data.
To capture detailed contact interactions during manipulation, we integrate the GelSight Mini sensor into a custom-built robotic gripper, enabling high-resolution tactile feedback at the fingertip.
From the sensor's raw tactile images, we extract estimates of the applied contact forces using \acf{feats} \cite{helmut2025learningforcedistributionestimation}. This force information is integrated as both observation and action in a diffusion policy, which is trained to replicate human demonstrations not only in terms of gripper motion, but also by explicitly predicting and controlling the target grip force applied to the object. This framework is designed to generalize across tasks requiring either strong or delicate manipulation. The following sections describe the gripper hardware, data collection and processing pipeline, the design of the \ac{farm} diffusion policy, and the implementation of closed-loop force control for deployment on a real robot.

\subsection{Gripper Hardware}
\label{sec:methods_gripper_hardware}
\begin{figure}[!t]
    \centering
    \includegraphics[trim={0pt 0pt 0pt 0pt}, clip, width=0.475\textwidth]{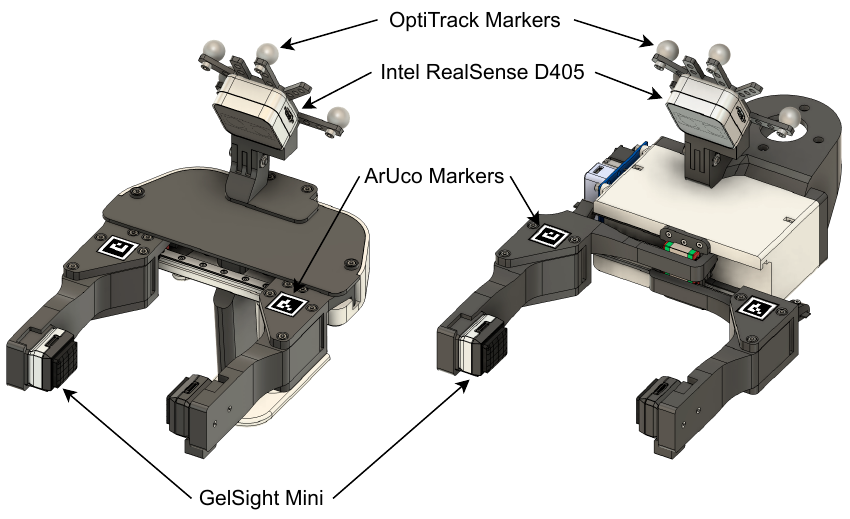}
    \caption{Side-by-side comparison of the adapted \ac{umi} gripper (left) used for demonstration data collection, and the Actuated UMI gripper (right) used for robotic deployment. Both designs feature an Intel RealSense D405 camera, a GelSight Mini tactile sensor (mounted on one fingertip), OptiTrack markers for motion tracking, and ArUco markers for grip width measurement. Sensor placement and overall geometry are matched to enable direct transfer of learned policies to the Actuated UMI.}
    \label{fig:hardware_setup_umi}
\end{figure}

Our approach relies on two closely related grippers: an adapted hand-held \acf{umi} gripper \cite{chi2024universal} for demonstration data collection, and a custom-built Actuated UMI gripper for robotic deployment (Fig.~\ref{fig:hardware_setup_umi}).
Our modified version of the \ac{umi} gripper replaces the original GoPro camera with an Intel RealSense D405, which provides in-hand RGB images and is tracked via OptiTrack markers for precise motion capture. The standard elastic TPU fingers are replaced with rigid fingers. One finger is fitted with a GelSight Mini sensor at its fingertip, and the other finger holds a shell of the GelSight Mini sensor with a matching gel pad but no electronics. In addition, we attach ArUco markers to each finger and use the in-hand Intel RealSense D405 camera to track their positions, enabling precise measurement of the grip width. 

To deploy learned policies on a real robot, we developed the Actuated UMI gripper. Powered by a single \mbox{DYNAMIXEL} XL430-W250-T motor, this gripper uses a belt-driven mechanism to synchronously actuate both fingers. Two coil springs, one per finger, return the gripper to its open position when torque is released. The geometry mirrors the adapted hand-held UMI gripper. All sensors and markers, including the GelSight Mini, the Intel RealSense D405, the OptiTrack markers, and the ArUco markers, are positioned identically, enabling direct transfer of policies learned on demonstration data. The Actuated UMI supports multiple control modes, such as position, velocity, pulse width modulation, and operates at approximately \SI{50}{\hertz}, making real-time force control feasible. We open-source all the design files and control software.

\subsection{Data Collection}
\label{sec:data_collection}
By using the adapted hand-held \ac{umi} gripper, we are able to both demonstrate the required motions and directly apply the necessary contact forces for each task. Capturing both gripper kinematics and high-resolution tactile feedback during demonstrations enables us to record the force profiles essential for learning policies that require precise grip control, something not achievable with conventional teleoperation or kinesthetic teaching methods. \cref{fig:data_collection} illustrates our demonstration setup along with the different sensor inputs.

\begin{figure}
    \centering
    \includegraphics[width=0.475\textwidth]{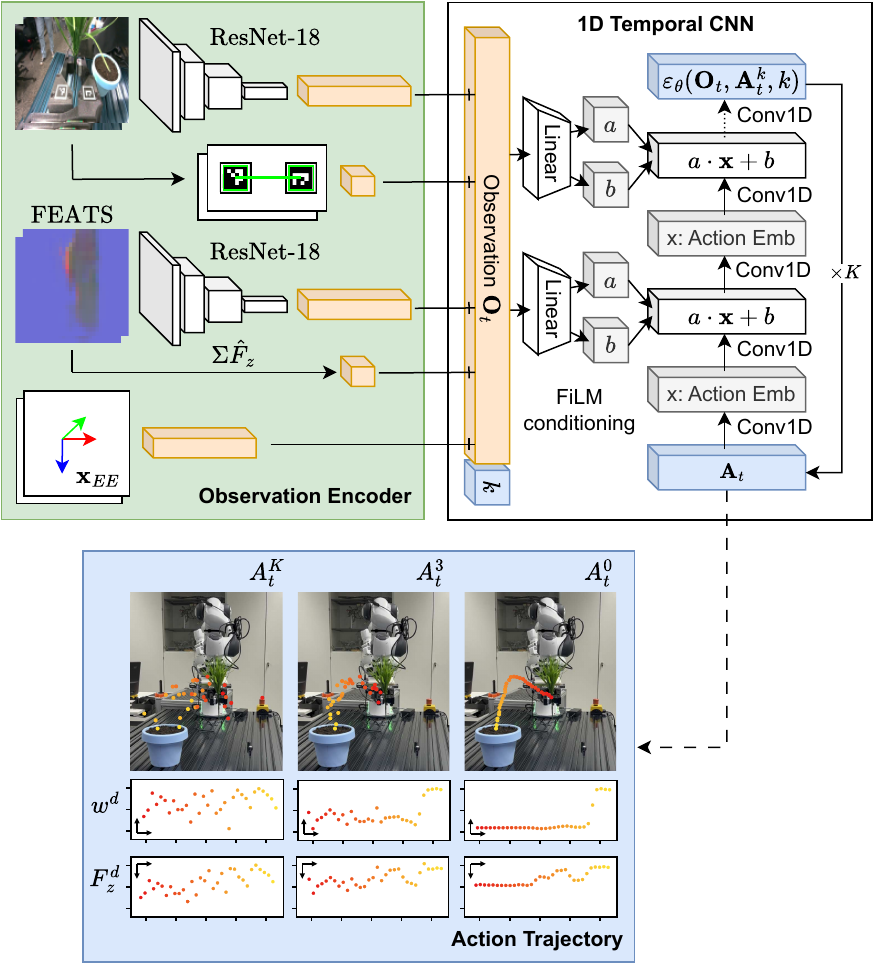}
    \caption[Schematic of the Force-Aware Diffusion Policy Architecture.]{Schematic of the \ac{farm} diffusion policy architecture. Visual, proprioceptive, and tactile observations are encoded and provided as input to a 1D temporal CNN with FiLM conditioning. The model predicts action trajectories including absolute end-effector pose, grip width, and grip force. This structure enables closed-loop force control of the gripper during manipulation.}
    \label{fig:farm_diffusion_policy}
\end{figure}

During each demonstration, we record synchronized streams of all relevant sensor data:
RGB images from the Intel RealSense D405 camera ($848 \times 480 \times 3$); grip width from the ArUco markers on the fingers, with positions determined by the in-hand Intel RealSense camera; gripper pose via OptiTrack, tracking the rigid body frame defined by the marker constellation; GelSight Mini tactile images ($320 \times 240 \times 3$); force distribution estimates computed from each GelSight Mini image using the open-source \ac{feats} model. 

\ac{feats}~\cite{helmut2025learningforcedistributionestimation} is a learning-based method that infers spatial force distributions from GelSight Mini images. It is trained on labeled data generated via finite element analysis. FEATS thus outputs physically grounded estimates of the shear and normal forces. We note that although FEATS demonstrates strong performance, deviations in the predictions may occur due to underlying model assumptions or challenges in generalizing across sensors.
Additionally, we represent the gripper pose through its 3D position and a 6D rotation representation~\cite{Zhou_2019_CVPR}.
All data is collected using \acf{ros}.
After recording, we synchronize all sensor streams to the GelSight Mini images, which operates at \SI{25}{\hertz} and act as the reference clock.
This ensures that every sample in the trajectory contains a complete set of sensor observations.

\subsection{FARM Diffusion Policy}
\label{sec:farm_diffusion_policy}
We build on the diffusion policy introduced by Chi et al.~\cite{chi2024diffusionpolicy}, using the 1D temporal \ac{cnn} variant with \ac{film} conditioning \cite{perez2017filmvisualreasoninggeneral}. To incorporate tactile feedback into the diffusion policy in a physically meaningful and interpretable way, we extend the observation and action spaces beyond those used in prior work \cite{chi2024diffusionpolicy}, \cite{chi2024universal}. Each policy input at time $t$ includes:

\begin{enumerate}
    \item \textbf{An in-hand RGB image} from the Intel RealSense D405 camera, downsampled to $96 \times 96 \times 3$ for computational efficiency without sacrificing the essential visual context.
    \item \textbf{Grip width}, calculated as the Euclidean distance between the centers of the ArUco markers on the two fingers, measured in the Intel RealSense D405 image.
    \item \textbf{Tactile feedback} is represented by a 3-channel image encoding the full force distributions extracted from each GelSight Mini tactile image using a pretrained and fixed \ac{feats} model. The image channels are (1) the shear force in the $x$-direction, (2) the shear force in the $y$-direction, and (3) the normal force in the $z$-direction, with each channel independently normalized and the overall image scaled to a $96 \times 96 \times 3$ resolution. Additionally, the total normal force is included as a scalar value, computed by integrating over the discretized normal force distribution, as it directly corresponds to the quantity regulated during closed-loop force control on the gripper.
    \item \textbf{The gripper pose}, consisting of 3D position coordinates and 6D rotation feature representation \cite{Zhou_2019_CVPR} providing an continuous encoding of spatial orientation.
\end{enumerate}

To process the visual inputs, we employ separate \mbox{ResNet-18} encoders for the in-hand RGB image and the tactile force image, enabling specialized feature extraction for each modality. The diffusion policy is designed to predict action trajectories consisting of the absolute target pose, target grip width $w^d$, and target grip force $F_z^d$, each over a fixed prediction horizon of $32$ and action execution horizon of $16$. The observation horizon comprises the two most recent observations, which allows the policy to capture short-term context for decision-making (Fig. \ref{fig:farm_diffusion_policy}). By incorporating both force and grip width as elements of the state and action space, these quantities are used for conditioning the model as well as being treated as output variables during the denoising process. This design ensures that the current grip force and grip width observations directly influence the predicted action trajectory, including the target force and target grip width at each step. As a result, the model's predictions align with current observations, mitigating the risk of implausible or unstable grasping behavior. This enables the policy to anticipate and regulate the amount of force required for subsequent steps.

The policy is trained using a mean squared error loss on the noise added during the denoising process. We use the publicly available implementation of the diffusion policy from LeRobot~\cite{cadene2024lerobot} with modifications to support our extended observation and action modalities.

\subsection{Policy Deployment and Gripper Control}
\label{sec:policy_deployment}
Including both target grip width and target force in the action space allows us to capture both positional and force-related aspects of manipulation. Target grip force is the relevant control variable during object contact because it enables closed-loop force control. However, target grip width is required to guide finger positioning during phases when there is no contact, such as when approaching, grasping, or releasing an object. Without explicit grip width actions, the policy would lack the means to open or close the gripper accurately outside the contact phase.

To deploy learned policies on the Actuated UMI gripper, we therefore implement a dual-mode control strategy that switches between grip width control and force control based on the current interaction phase. This strategy allows the robot to execute both pre-contact motions and in-contact, closed-loop force control. The diffusion policy outputs both a target grip width $w^d$ and a target force $F_z^d$ at each step. The controller monitors both the target force and the estimated contact force $\hat{F}_z$, computed from \ac{feats} using the latest GelSight Mini image $I_{GS}$. If both the target and estimated force are below \SI{-0.5}{\newton}, the system assumes that the robot is in contact with the object and switches to force control. Otherwise, the target grip width $w^d$ is directly sent to the internal PD controller of the DYNAMIXEL motor. The switching threshold of \SI{-0.5}{\newton} was selected based on the noise characteristics of the \ac{feats} model, ensuring that the controller only transitions to force control when actual contact is confidently detected. The controller computes the force error $e = \hat{F}_z - F_z^d$ and applies a PID controller to this error with anti-windup to ensure the integral term cannot accumulate beyond actuator limits.
This value is added to the current grip width and sent as a position command to the internal PD controller of the DYNAMIXEL motor.
The force control loop runs at \SI{25}{\hertz}, synchronized with GelSight Mini image acquisition and \ac{feats} prediction. During deployment, the diffusion policy runs at \SI{7}{\hertz}, while the lower-level force and position controllers operate at higher rates to bridge the gap between high-level action selection and real-time motor actuation.

To further ensure seamless transfer from demonstration data to robot execution, two calibrations are performed. First, hand-eye calibration \cite{tsai1989new} aligns the OptiTrack world frame with the robot base. This transformation allows us to directly command end-effector positions to the robot in its own base frame during policy execution. Second, a linear mapping between grip width, measured as the ArUco marker distance, and motor position is estimated via least squares by slowly closing the Actuated UMI gripper and recording both quantities. This ensures that grip width values predicted by the policy can be accurately converted into motor commands during deployment.

\subsection{Baselines}
To evaluate the contribution of tactile feedback and force control in our \ac{farm} framework, we compare it against three baseline strategies: (1) Force-Aware, (2) Tactile-Aware, and (3) Vision-Only.
Each baseline is also a diffusion policy and operates on a modified input representation and/or action space, while keeping the proprioceptive (end-effector pose, grip width) and visual (RGB image from Intel RealSense D405) inputs consistent with \cref{sec:farm_diffusion_policy}.

\subsubsection{Force-Aware Baseline}
This baseline uses the total normal force estimated by \ac{feats} from GelSight Mini images as the only tactile input, alongside vision, end-effector pose, and grip width. The action space includes target end-effector pose, target grip width, and target grip force, enabling closed-loop force control via the dual-mode controller described in \cref{sec:policy_deployment}. This baseline follows a similar pipeline as \ac{farm}, but omits the full force distribution, which limits its ability to capture detailed contact configurations.

\subsubsection{Tactile-Aware Baseline}
The tactile-aware baseline incorporates raw GelSight Mini images ($96 \times 96 \times 3$) as input, processed by a separate ResNet-18 encoder, alongside vision, end-effector pose, and grip width. The action space includes target end-effector pose and target grip width but excludes target force, preventing explicit force control. 
The reason for the lack of explicit force control is that while tactile feedback from raw GelSight Mini images provides implicit contact awareness, the policy cannot actively regulate contact forces based on having only the raw tactile images available.

\subsubsection{Vision-Only Baseline}
The vision-only baseline omits tactile feedback entirely, relying on vision, end-effector pose, and a binary gripper state (open/closed) derived from thresholding the grip width in demonstration data. The action space includes the target end-effector pose and a binary open/close command. During inference, the gripper moves to fully open or closed positions using the DYNAMIXEL motor's internal PD controller.

\section{Experiments \& Results}
\label{sec:experiments}
In this section, we evaluate our proposed \ac{farm} framework that combines physically-grounded high-dimensional tactile observations on the input level together with force-based grip control.
To investigate the contribution of both components, we compare the approach with the previously mentioned baselines, thereby assessing the role of tactile sensing and its representation for both the observation and action levels.
We structure the experiments along three main questions:  (1) \textit{How important is the explicit incorporation of force signals for task success?}
(2) \textit{How does the high-dimensional (image-based) force-distribution information impact task success?} and (3) \textit{How do these approaches compare in the force domain for the time-dependent task of screw tightening?}

\begin{figure}[t]
    \centering
    \subfloat[]{\includegraphics[height=9.1em]{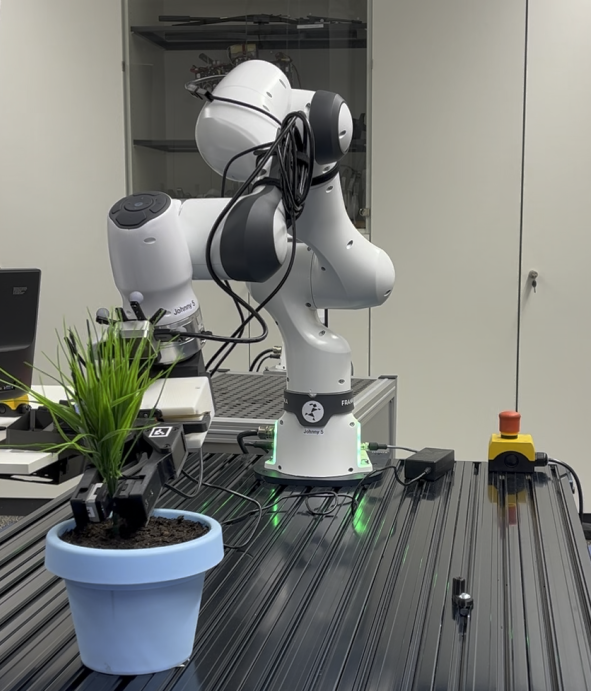}} \hskip 0.5em
    \subfloat[]{\includegraphics[height=9.1em]{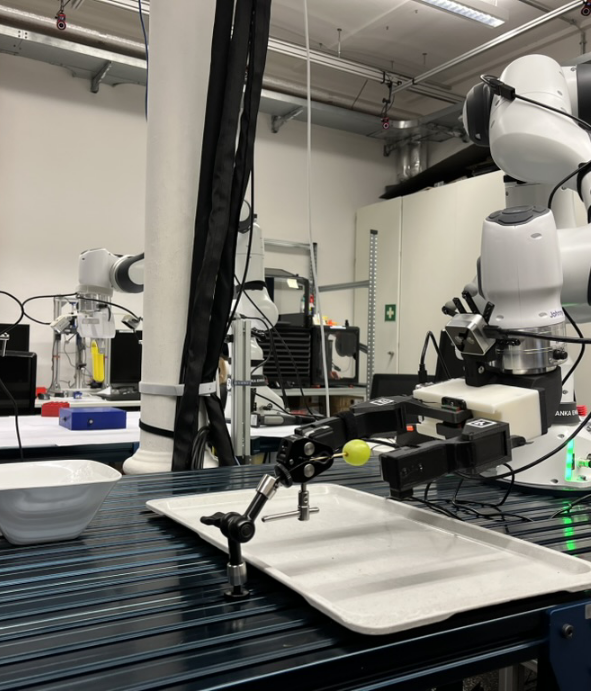}} \hskip 0.5em
    \subfloat[]{\includegraphics[height=9.1em]{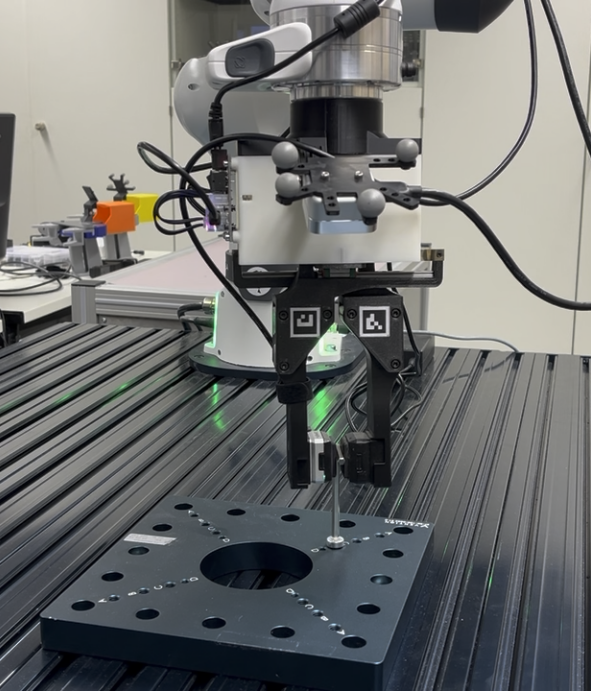}}
    \caption{Experiment setup for the three tasks: (a) plant insertion, showing the Franka Research 3 robot planting a plastic plant into a soil-filled pot; (b) grape picking, depicting the delicate grasping and detachment of a grape from a toothpick; (c) screw tightening, illustrating the robot using an Allen key to tighten a screw.}
    \label{fig:experiment_setup}
\end{figure}

\subsection{Experiment Setup} 
All experiments are conducted using a real Franka Research 3 robot equipped with the Actuated UMI gripper (see \cref{sec:methods_gripper_hardware}). 
The robot is controlled using the Cartesian position impedance controller from the franky control library \cite{Schneider_franky_High-Level_Control}.
Across all strategies, actuation of the Actuated UMI gripper relies on the internal PD position controller of its DYNAMIXEL motor. 

For all tasks, we collected $30$ demonstrations with our hand-held UMI gripper (see \cref{sec:data_collection}), all performed by the same expert.
We train the diffusion policies using \acf{ddpm} with $100$ denoising steps over $60000$ iterations. For inference, we adopt \acf{ddim} with $10$ steps to reduce the number of sampling iterations \cite{chi2024diffusionpolicy}.

The experiments involve three tasks, shown in \cref{fig:experiment_setup}, each testing different aspects of tactile-informed manipulation:

\noindent \textbf{Plant Insertion Task.} The task requires grasping a plastic plant from a fixed holder, inserting it into a flower pot filled with moistened soil, and finally releasing it. This task requires a high grip force to ensure stable insertion without losing grasp. A rollout is considered successful if the plant remains upright in the soil without tilting or touching the pot rim.
To standardize the initial conditions across all approaches, we constrain the rollout until grasp detection.

\noindent \textbf{Grape Picking Task.} This task consists of grasping a grape from a toothpick without crushing or letting the grape slip from the robot's grip. Successful execution here consists of grasping the grape, delicately removing it from the toothpick, and finally, placing the grape into a bowl. A rollout is successful if the grape is placed intact and visually undamaged into the bowl.

\noindent \textbf{Screw Tightening Task.} This task requires grasping an Allen key inserted in a screw at a fixed position, rotating it, and opening the gripper upon having tightened the screw. 
To introduce additional variability into the task, we consider two relative orientations between the Allen key and the screw, resulting in variation in the final tightened position. 
Success is achieved if the screw is noticeably tight.
This task relies on tactile feedback to detect tightness, as vision alone cannot determine the screw's state. For safety, rollouts are terminated if the end-effector torque exceeds a threshold.

\subsection{Results}

\begin{figure}
    \centering
    \includegraphics[width=\linewidth]{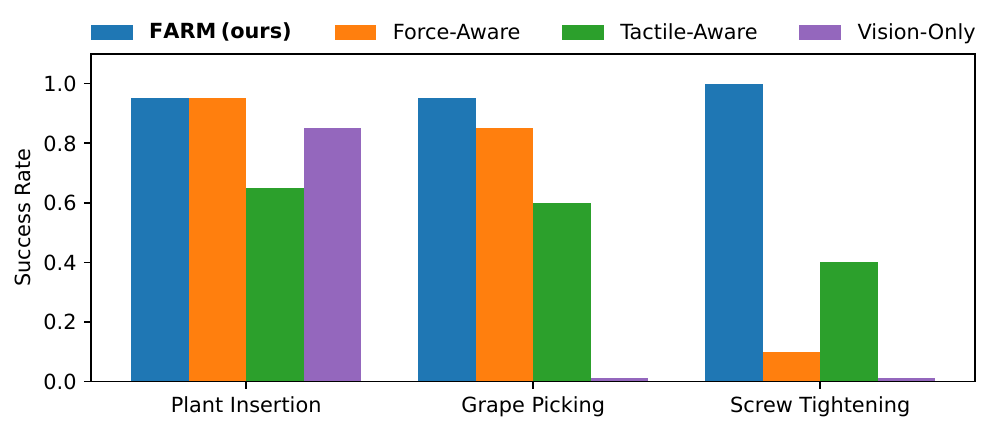}
    \caption{Success rates for the plant insertion, grape picking, and screw tightening tasks, comparing the \ac{farm} framework against force-aware, tactile-aware, and vision-only baselines, evaluated over 20 rollouts per task.}
    \label{fig:success_rates}
\end{figure}

\subsubsection{How important is the explicit incorporation of force signals for task success?}
To evaluate the importance of explicit force signals over raw tactile images or vision-only inputs, we compare the force-aware, tactile-aware, and vision-only baselines across all tasks. Success rates are assessed over $20$ rollouts per task, with results summarized in \cref{fig:success_rates}.

In the plant insertion task, the force-aware baseline achieves a $95\%$ success rate, surpassing tactile-aware, with $65\%$ success rate, and the vision-only baseline with $85\%$ success. The explicit force control in the force-aware baseline ensures sufficient grip to prevent slippage during plant transport and insertion into soil. Tactile-aware baseline, lacking direct force modulation, frequently applies insufficient grasp force, leading to slippage or incomplete planting. The vision-only baseline performs moderately well due to its binary gripper command by applying high grasping forces, but its lack of tactile feedback limits its ability to confirm stable grasps and successful plant insertion, resulting in occasional failures by opening the gripper too early. 

In the grape picking task, the force-aware baseline achieves $85\%$ success rate, compared to the tactile-aware baseline's $60\%$ success and vision-only $0\%$ success. The force-aware baseline's closed-loop force control applies precise force to avoid crushing the grape while maintaining a secure grip. With the tactile-aware baseline's passive use of raw tactile images, it often results in weak grips, causing slippage. The vision-only baseline fails entirely, as its coarse binary gripper command consistently crushes the grape. 

Finally, in the screw-tightening task, where tactile feedback is essential for detecting screw tightness, the force-aware baseline achieves  $10\%$ success rate and the tactile-aware baseline $40\%$. Both surpass the vision-only baseline, which once again fails entirely. In this case, due to a lack of tactile feedback needed to sense screw resistance, it consistently opens the gripper prematurely. The scalar force signal available to the force-aware baseline enables partial detection of this resistance but is insufficient for maintaining proper Allen key alignment, often causing it to slip out of the screw head. In contrast, the tactile-aware baseline leverages higher-dimensional information to outperform the force-aware baseline. However, this baseline also frequently opens the gripper too early, missing tightness cues due to its lack of explicit force measurements.

These results highlight that explicit force incorporation, as in the force-aware baseline, improves task success in manipulation tasks that require consistent static high or low forces, such as plant insertion and grape picking. However, its limited performance in the screw tightening task indicates that scalar force signals alone may be insufficient for tasks requiring precise contact state detection, suggesting the need for richer force representations. To address this, in the next sections, we investigate how incorporating the full force distribution information impacts task success.

\subsubsection{How does the high-dimensional (image-based) force-distribution information impact task success?}
Building on the established importance of explicit force signals, we now examine how high-dimensional, image-based force-distribution information, as used in the \ac{farm} framework, enhances task success compared to scalar force signals, raw tactile images, and no tactile feedback. \ac{farm} encodes shear and normal forces as images (\cref{fig:farm_diffusion_policy}), providing richer contact information than the scalar normal force in the force-aware baseline or the raw tactile image in the tactile-aware baseline. 

As seen in \cref{fig:success_rates}, in the plant insertion and grape picking tasks, \ac{farm} achieves success rates of $95\%$ for both tasks, which is comparable to the force-aware baseline with $95\%$ and $85\%$ respectively. These results indicate that scalar force signals suffice for tasks with consistent static force demands. The screw tightening task best highlights the advantages of force distributions over scalar force signals or raw tactile images, with \ac{farm} achieving a $100\%$ success rate compared to force-aware $10\%$, tactile-aware $40\%$, and vision-only $0\%$ baselines. \ac{farm}'s tactile force images capture shear and normal force interactions, enabling both a more precise downward pressure to keep the Allen key engaged and the accurate detection of screw tightness. In comparison, the force-aware baseline relies only on a scalar normal force signal, struggling to maintain proper Allan key alignment and often lifting it out of the screw head. The tactile-aware, while better than the force-aware policy, still fails to reliably interpret the complex contact dynamics from raw tactile images, leading to premature gripper opening. The vision-only baseline lacks the needed tactile feedback for this task.

These results show that high-dimensional force-distribution information significantly enhances task success, particularly in tasks requiring nuanced force interactions. By capturing both shear and normal forces, \ac{farm} provides superior control and contact state awareness compared to scalar force signals or raw tactile images. This advantage is most pronounced in tasks where temporal and multi-dimensional force dynamics are critical.

\subsubsection{How do these approaches compare in the force domain for the time-dependent task of screw tightening?}

\begin{figure}[t]
    \centering
    \vspace{-2mm}
    \subfloat{\includegraphics[width=0.49\linewidth]{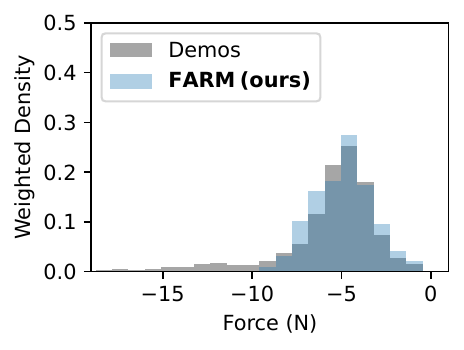}}
    \subfloat{\includegraphics[width=0.49\linewidth]{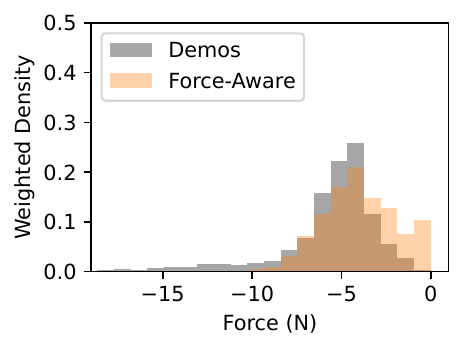}} 
    \vspace{-5mm}
    \subfloat{\includegraphics[width=0.49\linewidth]{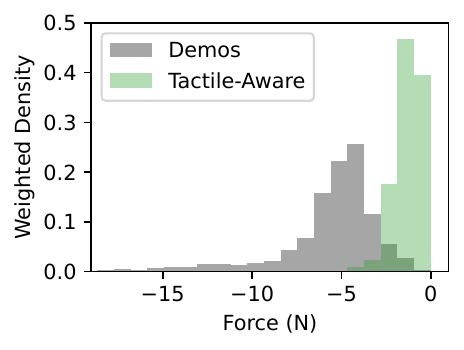}}
    \subfloat{\includegraphics[width=0.49\linewidth]{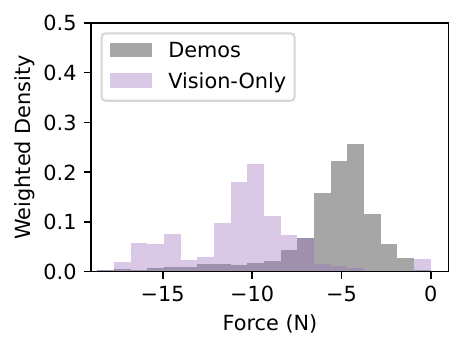}}
    \caption{Weighted empirical force distributions for the screw tightening task, comparing demonstrations and rollouts for the \ac{farm} framework against force-aware, tactile-aware, and vision-only baselines.}
    \label{fig:wasserstein_distance}
\end{figure}

To answer this question, we compare \ac{farm} in the force domain against the force-aware, tactile-aware, and vision-only baselines in the time-dependent screw tightening task. To quantify how closely the applied forces from each approach match those of the demonstrations, we use the Wasserstein-1 distance $W_1$ \cite{villani2021topics}.
Since trajectories differ in duration, simply pooling all samples would bias the statistics toward longer trajectories. For that reason, we use an equal-mass weighting scheme, ensuring that each demonstration or policy rollout contributes equally regardless of trajectory length. For trajectory $m$ with $n_m$ samples, each sample receives the weight $w_{k,m} = 1 / {M \cdot n_m}$, where $M$ is the number of trajectories. The $W_1$ distance is then defined as

\begin{equation}
W_1(u,v) = \inf_{\pi \in \Gamma(u,v)} \int_{\mathbb{R} \times \mathbb{R}} |x - y| d\pi(x,y),
\end{equation}
for weighted empirical force distributions of demonstrations $u$ and policy rollouts $v$. The $W_1$ value corresponds to the average amount the force distributions from one set would need to be shifted to match the other, expressed in the same unit as the force~(N). Smaller values, therefore, indicate that the forces measured during rollouts are more similar to those of the demonstrations, while larger values indicate greater differences in either magnitude or distributional shape.

As shown in \cref{fig:wasserstein_distance}, \ac{farm}'s weighted empirical force distributions closely align with those of the demonstrations. This is also represented by the Wasserstein-1 distance of \SI{0.7538}{\newton}. The force-aware baseline follows with a distance of \SI{1.6580}{\newton}, showing reasonable, but less precise force matching. Although the force-aware baseline was able to produce similar forces, it nevertheless failed in the task success because it lacked the necessary contact state information from just a single force value. The tactile-aware baseline, with a Wasserstein-1 distance of \SI{4.3801}{\newton}, tends to apply insufficient forces, while the vision-only baseline, at \SI{5.0515}{\newton}, consistently applies excessive forces, leading to significant deviations from the demonstration force profiles. These results show that incorporating force distributions into the observation embedding enables \ac{farm} to outperform the baselines in the force domain. The larger $W_1$ distances for force-aware, tactile-aware, and vision-only baselines reflect their limitations in capturing the complex, time-varying force profiles required for precise screw tightening.

\section{Conclusion}
\label{sec:conclusion}
This work investigated how tactile sensing can be incorporated not only as an observation modality but also directly into the action space of imitation learning policies for contact-rich manipulation. 
We thus introduced \ac{farm}, a visuotactile-conditioned diffusion policy that predicts target grip width and target grip force jointly with the robot pose.
Within \ac{farm}, the tactile-conditioned contact information is used both as an observation and as a basis for generating force-based action sequences.
Demonstrations were collected using an adapted hand-held \ac{umi} gripper equipped with a GelSight Mini sensor at one fingertip, with normal and shear force distributions estimated via the \ac{feats} model.
To enable policy transfer, we developed the \emph{Actuated UMI} gripper, whose geometry and kinematics match the hand-held \ac{umi} gripper.
We executed gripper actions through a dual-mode controller that switches between grip width position control and closed-loop force control.
The real-world experimental results and comparison with several baselines show the effectiveness of our proposed method and underline the importance of both leveraging a force-based action space, as well as physically-grounded, high-dimensional tactile information as an observation modality.
In addition to improved reliability and success rates, our proposed method also aligns more closely with the demonstration data w.r.t. the applied forces.
Future work should explore generalizing the method to bimanual manipulation, incorporating anthropomorphic hands, and employing a flow-matching objective to enhance the reactivity of the learned policies.

\typeout{}
\bibliographystyle{IEEEtran}
\bibliography{references}

\end{document}